\def\year{2019}\relax
\documentclass[letterpaper]{article} 
\usepackage{aaai18}  
\usepackage{times}  
\usepackage{helvet}  
\usepackage{courier}  
\usepackage{url}  
\usepackage{graphicx}  
\usepackage[utf8]{inputenc} 
\usepackage[T1]{fontenc}    
\usepackage{hyperref}       
\usepackage{booktabs}       
\usepackage{amsfonts}       
\usepackage{nicefrac}       
\usepackage{microtype}      
\usepackage{graphicx}
\usepackage{float}
\usepackage{comment}
\usepackage{amsmath}
\usepackage{algorithm}
\usepackage{algpseudocode}
\usepackage{subcaption}
\usepackage{soul}
\newcommand{\citet}[1]
{\citeauthor{#1}~\shortcite{#1}}
\newcommand{\citep}{\cite}

\begin{document}
%
\title{Community Regularization of Visually-Grounded Dialog}
\author{Akshat Agarwal*, Swaminathan Gurumurthy*, Vasu Sharma*\\
School of Computer Science\\
Carnegie Mellon University\\
\texttt{aa7@cmu.edu,sgurumur@andrew.cmu.edu,vasus@andrew.cmu.edu}
\AND Mike Lewis\\
School of Computing and Information\\
University of Pittsburgh\\
\texttt{ml@sis.pitt.edu}\\
\And Katia Sycara\\
School of Computer Science\\
Carnegie Mellon University\\
\texttt{katia@cs.cmu.edu}
}
\maketitle
\newcommand\ak[1]{{\color{}{#1}}}
\newcommand\gs[1]{{\color{}{#1}}}
\newcommand\akr[1]{{\color{}{#1}}}
\newcommand\gsr[1]{{\color{}{#1}}}

\begin{abstract}
The task of conducting visually grounded dialog involves learning goal-oriented cooperative dialog between autonomous agents who exchange information about a scene through several rounds of questions and answers in natural language. We posit that requiring artificial agents to adhere to the rules of human language, while also requiring them to maximize information exchange through dialog is an ill-posed problem. We observe that humans do not stray from a common language because they are social creatures who live in communities, and have to communicate with many people everyday, so it is far easier to stick to a common language even at the cost of some efficiency loss. Using this as inspiration, we propose and evaluate a multi-agent community-based dialog framework where each agent interacts with, and learns from, multiple agents, and show that this community-enforced regularization results in more relevant and coherent dialog (as judged by human evaluators) without sacrificing task performance (as judged by quantitative metrics).
\end{abstract}

\section{Introduction}
Intelligent assistants like Siri and Alexa are increasingly becoming an important part of our daily lives, be it in the household, the workplace or in public places. As these systems become more advanced, we will have them interacting with each other to achieve a particular goal \cite{leviathan2018}. We want these conversations to be interpretable to humans for the sake of transparency and ease of debugging. Having the agents communicate in natural language is one of the most universal ways of ensuring interpretability. This motivates our work on goal-driven agents which interact in coherent language understandable to humans. \gs{Most prior work on visual dialog \cite{visdial,vdialog_old} has approached the problem using supervised learning where}, conditioned on the question - answer pair dialog history, a caption $c$ and the image $I$, the agent is required to answer a given question $q$. The model is trained in a supervised learning framework using ground truth supervision from a human-human dialog dataset. 

\begin{figure}
    \centering
    \includegraphics[width=0.48\textwidth]{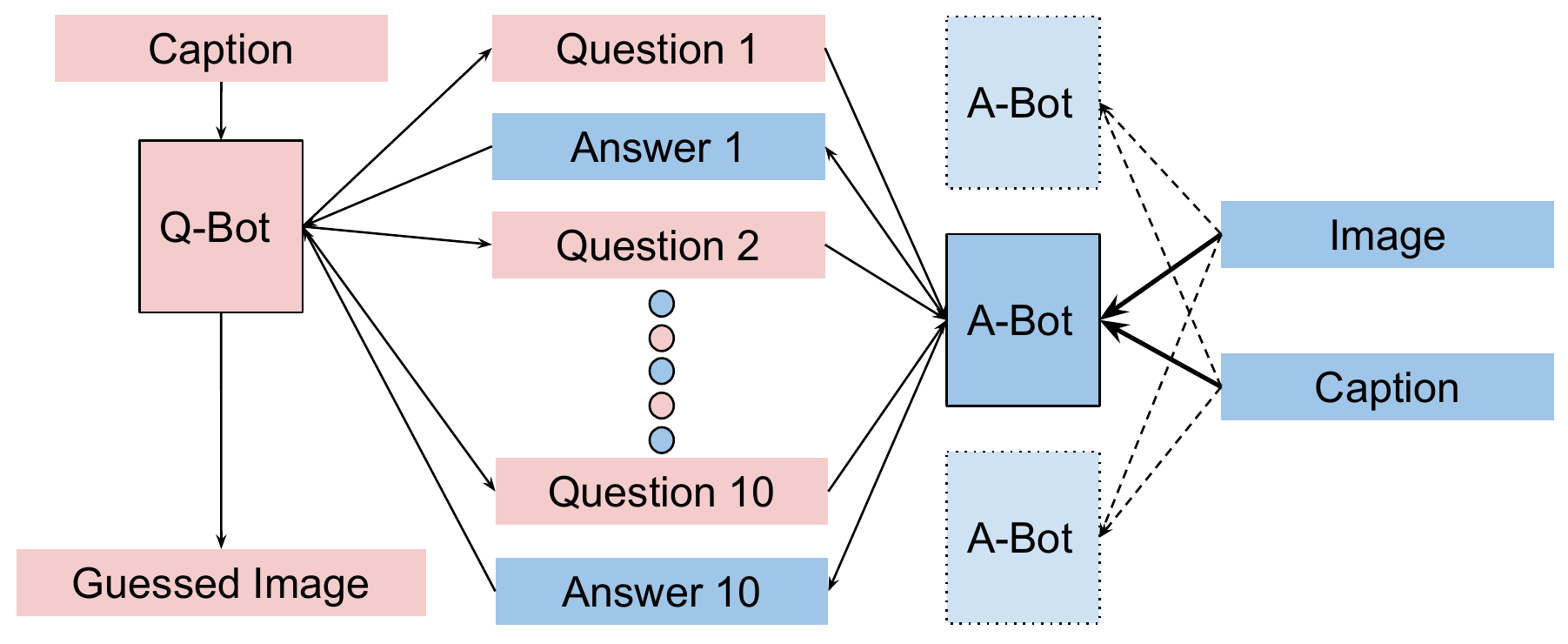}
    \caption{Multi-Agent (with 1 Q-Bot, 3 A-Bots) Dialog Framework}
    \label{fig:diagram}
\end{figure}

\gs{Some recent work \cite{vdialog} has tried to approach the problem using reinforcement learning, with two agents, namely the Question (Q-) Bot and the Answer (A-) Bot. }While the A-Bot still has the image, caption and the dialog history to answer any question, the Question Bot only has access to the caption and the dialog history. \ak{The two agents are initially trained with supervision using the VisDial v0.9 dataset \cite{visdial}, which consists of 80k images, each with a caption and 10 human generated question-answer pairs discussing the image. Under supervision, the agents are trained in an isolated manner to maximize the likelihood of generating the ground truth answers.\gs{ The agents are then made to interact and talk to each other, with a common goal of trying to improve the Q-Bot's understanding of the image.} The agents learn from their conversation with each other via reinforcement learning. While the supervised training \textit{in isolation} helps the agents to learn to interpret the images and communicate information, it is the \textit{interactive} training phase which leads to richer dialog with more informative questions and answers as the agents learn to adapt to each others' strengths and weaknesses. However, it is important to note that the optimization problem in this conversational setting does not make the agents stick to the domain of grammatically correct and coherent natural language.\gs{ Indeed, if the two agents are allowed to communicate and learn from each other for too long, they quickly start generating non-grammatical and semantically meaningless sentences}. While the generated sentences stop making sense\gs{ to human observers, the two agents are able to understand each other much better, and the Q-Bot's understanding of the image improves. This is similar to how twins often develop a private language \cite{rutter2003twins}, an idiosyncratic and} exclusive form of communication understandable only to them. This, however, reduces transparency of the agents' dialog to any observer (human or AI), and is hence undesirable. Prior work \cite{visdial,vdialog} which has focused on improving  performance as measured by the Q-Bot's image retrieval rank has suffered from incoherent dialog. We address this problem of improving the agents' performance while increasing dialog quality by taking inspiration from humans. We observe that humans continue to speak in commonly spoken languages, \akr{and hypothesize that this is} \textit{because they need to communicate with an entire community}, \akr{and having a private language for each person would be extremely inefficient}. With this idea, we let our agents learn in a similar setting, by making them talk to (ask questions of, get answers from) multiple agents, one by one.We call this \textit{Community Regularization}.

In the subsequent sections we describe the Visual Dialog task and the neural network architectures of our Q-Bots and A-Bots in detail. We then describe the training process of the agents sequentially: (a) in isolation (via supervision), (b) while interacting with one partner agent (via reinforcement), and (c) our proposed multi-agent setup where each agent interacts with multiple other agents (via reinforcement). We compare the performance of the agents trained in these different settings, both quantitatively using image retrieval ranks, and qualitatively evaluating the overall coherence, grammar and relevance of the dialog generated, as judged by impartial human evaluators. We make the following contributions:  we show that community regularization resulting from our multi-agent dialog setup ensures that the interactions between the agents remain grounded in the rules and grammar of natural language, are coherent and human-interpretable \akr{without compromising on task performance}}. We make our code available as open-source\footnote{\url{https://github.com/agakshat/visualdialog-pytorch}}.

\section{Problem Statement}
We begin by defining the problem of Visually Grounded Dialog for the co-operative image guessing game on the VisDial dataset. 

\textbf{Players and Roles}: The game involves two collaborative agents – a question bot (Q-bot) and an answer bot (A-bot). \ak{The A-bot has access to an image and caption, while the Q-bot has access to the image's caption, but not the image itself. Both the agents share a common objective, which is for the Q-bot to form a good ``mental representation" of the unseen image which can be used to retrieve, rank or generate that image. This is facilitated by the exchange of 10 pairs of questions and answers between the two agents, using a shared vocabulary, where the Q-bot asks the A-bot a question about the image, and the A-bot answers the question, hence improving the Q-Bot's understanding of the image scene.}

\textbf{General Game Objective}: At each round, in addition to communicating with the A-bot, the Q-bot also provides the learning algorithm with its best estimate $y_t$ of the unknown image $I$ based only on the dialog history and caption. Both agents receive a common reward from the environment which is inversely proportional to the error in this description under some metric $L(y_t, y_{gt})$. We note that this is a general setting where the `description’ $y_t$ can take on varying levels of specificity – from \ak{image feature embeddings extracted by deep neural networks to textual descriptions and pixel-level image re-generations.}

\begin{figure*}
    \centering
    \includegraphics[height=0.35\textwidth]{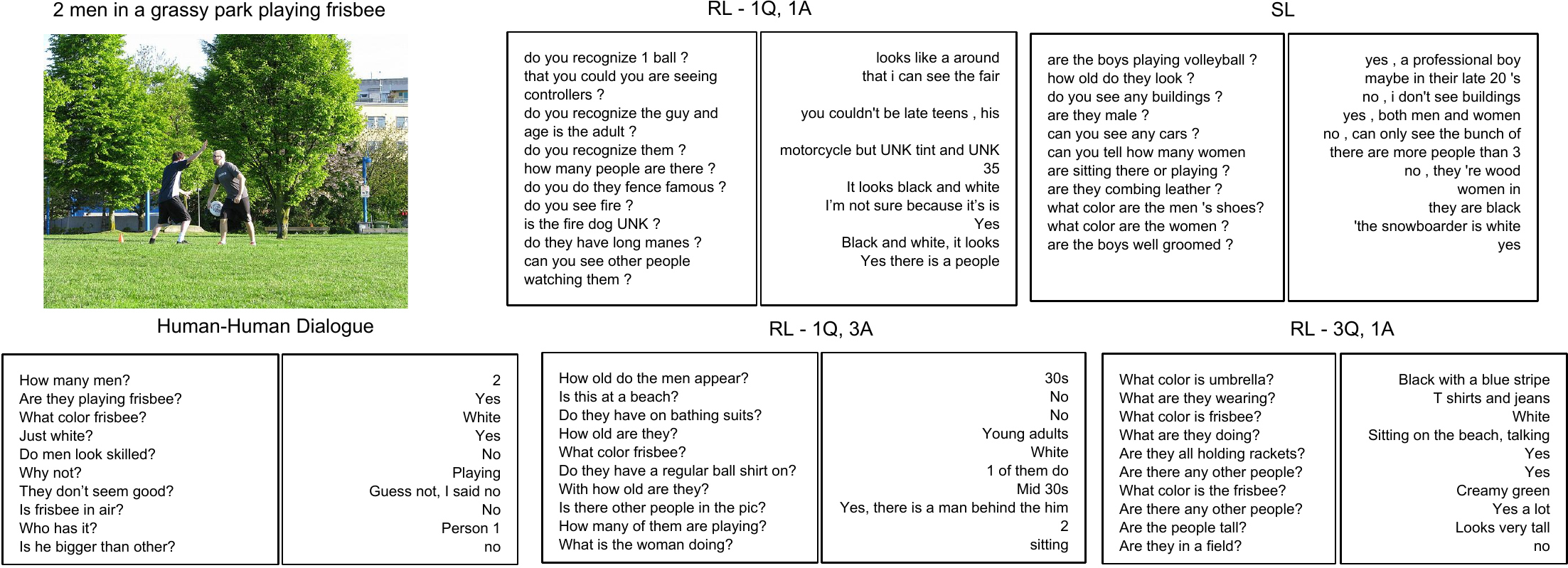}
    \caption{A randomly selected image from the VisDial dataset followed by the ground truth (human) and generated dialog about that image for each of our 4 systems (SL, RL-1Q,1A, RL-1Q,3A, RL-3Q,1A). This example was one of 102 used in the human evaluation results shown in Table \ref{table:humaneval}}
    \label{fig:humaneval}
\end{figure*}

\textbf{Specific Instantiation}: \gs{In our experiments, we focus on the setting where the Q-bot is tasked with estimating a vector embedding of the image I, which is later used to retrieve a similar image from the dataset}. \ak{Given a feature extractor (say, a pretrained CNN model like VGG \cite{vgg}), the target `description’ $y_{gt}$ of the image, can be obtained by simply forward propagating through the VGG model, without the requirement of any human annotation. Reward/error can be measured by the Euclidean distance between the target description $y_{gt}$ and the predicted description $y_t$, and any image may be used as the visual grounding for a dialog. Thus, an unlimited number of games may be simulated without human supervision, motivating the use of reinforcement learning in this framework.

Our primary focus for this work is to ensure that the agents' dialog remains coherent and understandable while also being informative and improving task performance. For concreteness, consider an example of dialog that is informative yet incoherent: \textbf{question}: "do you recognize the guy and age is the adult?", \textbf{answered with}: "you couldn't be late teens, his".} \gs{The example shows that the bots try to extract and convey as much information as possible in a single question/answer (sometimes by incorporating multiple questions or answers into a single statement). But in doing so they lose basic semantic and syntactic structure. We also provide a sample of the dialogs in Figure \ref{fig:humaneval}.}

\section{Related Work}
Most of the major works which combine vision and language have traditionally focused on the problem of image captioning ((\cite{kiros}, \cite{caps1}, \cite{caps2}, \cite{caps3}, \cite{caps4}, \cite{caps5}) and visual question answering (\cite{vqa}, \cite{vqa2}, \cite{vqa3}).
The problem of visual dialog is relatively new and was first introduced by \citet{visdial} who also created the VisDial dataset to advance the research on visually grounded dialog. The dataset was collected by pairing two annotators on Amazon Mechanical Turk to chat about an image. They formulated the task as a `multi-round' VQA task and evaluated individual responses at each round in an image guessing setup. In subsequent work by \citet{vdialog} they proposed a reinforcement learning based setup where they allowed the Question bot and the Answer bot to have a dialog with each other with the goal of correctly predicting the image unseen to the Question bot. However, in their work they noticed that the reinforcement learning based training quickly led the bots to diverge from natural language. In fact \citet{NLemergence} recently showed that language emerging from two agents interacting with each other might not even be interpretable or compositional. We use community regularization to alleviate this problem.
Recent work has also proposed using such goal driven dialog agents for other related tasks including negotiation \cite{deal} and collaborative drawing \cite{codraw}. We believe that our work can easily extend to those settings as well.
\citet{vdialog2} proposed a generative-discriminative framework for visual dialog where they train only an answer bot to generate informative answers for ground truth questions. These answers were then fed to a discriminator, which was trained to rank the generated answer among a set of candidate answers. This is a major restriction of their model as it can only be trained when this additional information of candidate answers is available, which restricts it to a supervised setting. Furthermore, since they train only the answer bot and have no question bot, they cannot simulate an entire dialog which also prevents them from learning by self-play via reinforcement. \citet{vdialog3} further improved upon this generative-discriminative framework by formulating the discriminator as a more traditional GAN \cite{gan}, where the adversarial discriminator is tasked to distinguish between human generated and machine generated dialogs. 


\section{Agent Architectures}
\label{sec:arch}
We briefly describe the agent architectures in this section and leave the details for the appendix.

\subsection{Question Bot Architecture}
\begin{figure}[ht]
\centering
\begin{subfigure}{0.46\textwidth}
\centering
\includegraphics[width=1\textwidth]{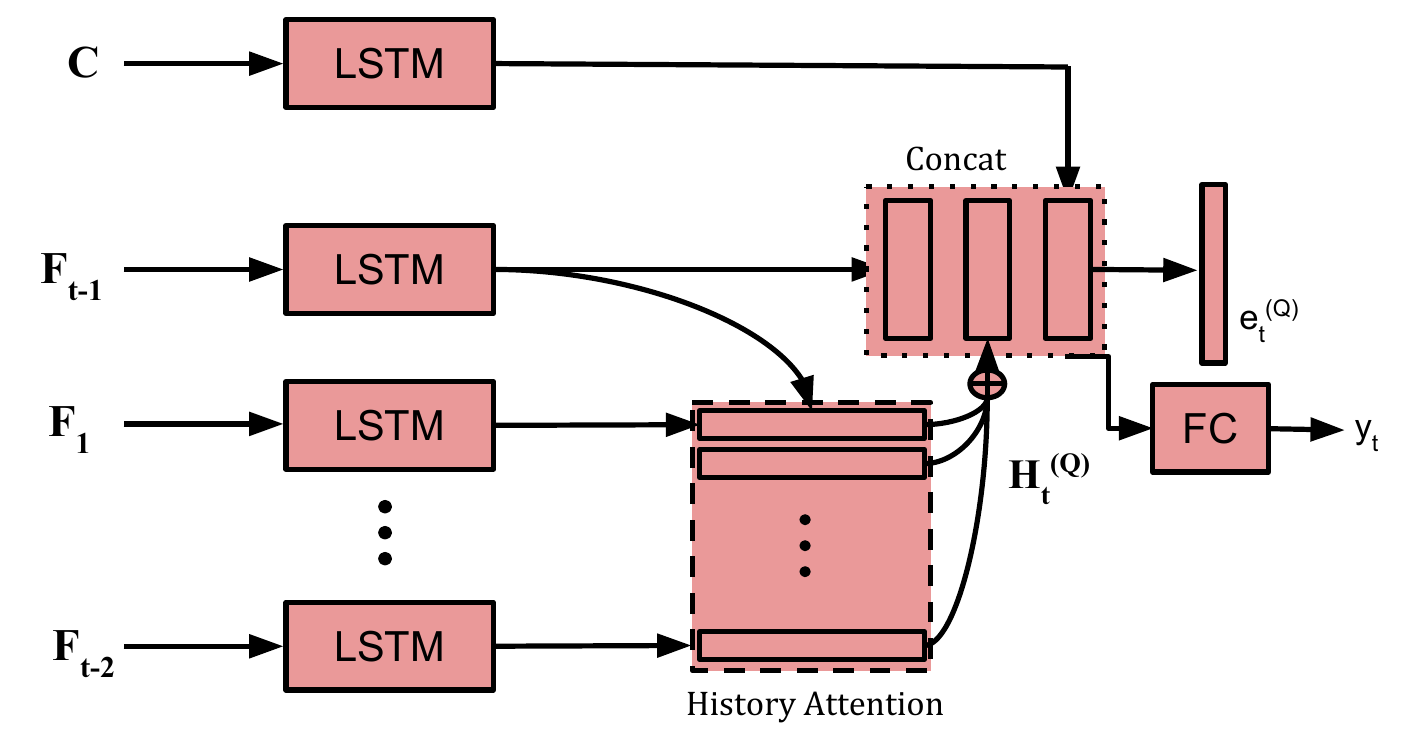}
\caption{Encoder architecture for Q-Bot}
\label{im:im2}
\end{subfigure}

\begin{subfigure}{0.46\textwidth}
\centering
\includegraphics[width=1\textwidth]{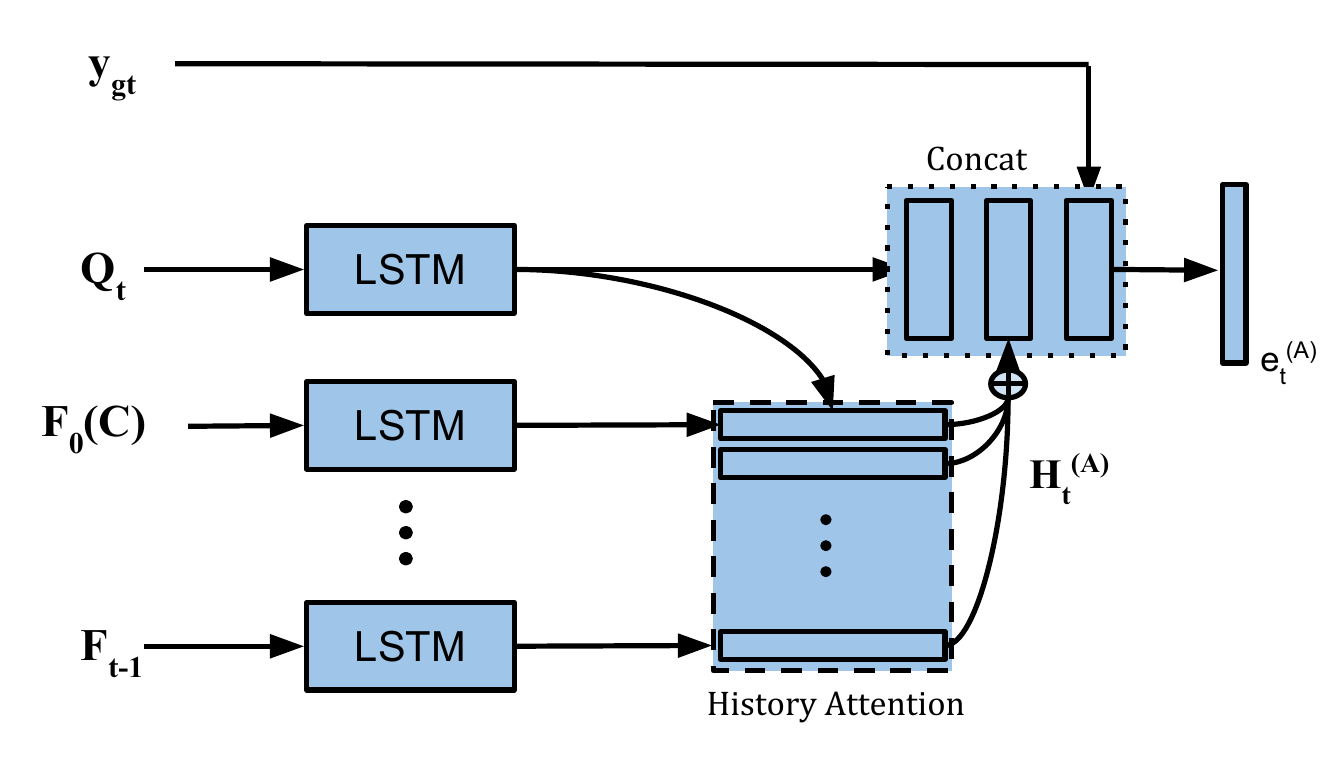}
\caption{Encoder architecture for A-Bot}
\label{im:im1}
\end{subfigure}
\label{fig:encoders}
\caption{}
\end{figure}

\ak{\gs{The question bot architecture we use is inspired by the answer bot architecture in \citet{vdialog,vdialog2} but the individual units have been modified to provide more useful representations. Similar to the original architecture, our Q-Bot, shown in Fig. \ref{im:im2}, also consists of 4 parts, (a) fact encoder, (b) state-history encoder, (c) question decoder and (d) image regression network. The fact encoder is modelled using a Long-Short Term Memory (LSTM) network, which encodes the previous question-answer pair into a fact embedding $F_t$. We modify the state-history encoder to incorporate a two-level hierarchical encoding of the dialog. It uses the fact embedding $F_t$ at each time step to compute attention over the history of dialog, $(F_1, F_2, F_3... F_{t-1})$ and produce a history encoding $H_t^Q$. The key modification (compared to \citet{vdialog2}) in our framework is the addition of a separate LSTM to compute a caption embedding $C$. This is key to ensuring that the hierarchical encoding does not exclusively attend on the caption while generating the history embedding. }The caption embedding is then concatenated with $F_t$ and $H_t^Q$, to obtain $S_t^Q$. $S_t^Q$ is then passed through multiple fully connected layers to compute the state-history encoder embedding $e_t^Q$ and the predicted image feature embedding $y_t = f(S_{t}^Q)$.} The encoder embedding, $e_t^Q$ is fed to the question decoder (not pictured), another LSTM, which generates the question, $q_t$. For all LSTMs and fully connected layers in the model we use a hidden layer size of 512. The image feature vector is 4096 dimensional. The word embeddings and the encoder embeddings are 300 dimensional. 

\subsection{Answer Bot Architecture}
\gs{The architecture for A-Bot, also inspired from \citet{vdialog2}, shown in Fig. \ref{im:im1}, is similar to that of the Q-Bot. It has 3 components: (a) question encoder, (b) state-history encoder and (c) answer decoder.} The question encoder computes an embedding, $Q_t$ for the question to be answered, $q_t$. The history encoding $(F_1, F_2, F_3... F_{t}) \rightarrow  H_t^A$ uses a similar two-level hierarchical encoder, where the attention is computed using the question embedding $Q_t$. The caption is passed on to the A-Bot as the first element of the history, which is why we do not use a separate caption encoder. Instead, we use the fc7 feature embedding of a pretrained VGG-16 \cite{vgg} model to compute the image embedding $I$. The three embeddings $H_t^A, Q_t, y_{gt}$ are concatenated and passed through another fully connected layer to extract the encoder embedding $e_t^A$. The answer decoder (not visualized), which is another LSTM, uses this embedding $e_t^A$ to generate the answer $a_t$. Similar to the Q-Bot, we use a hidden layer size of 512 for all LSTMs and fully connected layers. The image feature vector coming from the CNN is 4096 dimensional (FC7 features from VGG16). The word embeddings and the encoder embeddings are 300 dimensional. 

\section{Training}
\label{sec:training}
\gs{We follow the training process proposed in \citet{vdialog}. Two agents, a Q-Bot and an A-Bot are first trained in isolation, by supervision from the VisDial dataset. After this supervised pretraining for 15 epochs over the data, we smoothly transition the agents to learn from each other via reinforcement learning. The individual phases of training will be described in more detail below with some details in the appendix.}

\subsection{Supervised pre-training}
\label{sec:supervised}
In the supervised part of training, both the Q-Bot and A-Bot are trained separately, using a \ak{Maximum Likelihood Estimation (MLE) loss computed using the ground truth questions and answers, respectively, for every round of dialog. The Q-Bot simultaneously optimizes another objective: minimizing the Mean Squared Error (MSE) loss between the true ($y_{gt}$) and predicted ($y_t$) image embeddings. The ground truth dialogs and image embeddings are from the VisDial dataset.}

Given the true dialog history, image features and a question from the dataset, the A-Bot generates an answer to that question. Given the true dialog history and the previous question-answer pair from the dataset, the Q-Bot is made to generate the next question to ask the A-Bot. Both agents receive only ground truth questions and answers, never what the other agent generated - so the two agents never actually interact during this phase of training. \gs{However, there are multiple problems with this training scheme. First, MLE is known to result in models that generate repetitive dialogs and often produce generic responses. Second, since the agents are never allowed to interact during training, they end up facing out of distribution questions and answers when made to interact during evaluation, \gs{which reduces the task performance. This can be observed in Figure \ref{fig:percentile}. The performance of the agents trained via supervised learning dips after each successive dialog round.}}

\subsection{Reinforcement Learning Setup}
\label{sec:rl}
To alleviate the issues pointed out with supervised training, we let the two bots interact with each other via self-play (no ground-truth except images and captions). This interaction is also in the form of questions asked by the Q-Bot, and answered in turn by the A-Bot, using a shared vocabulary. The state space is partially observed and asymmetric, with the Q-Bot observing $\{c,q_{1},a_{1}\ldots q_{10},a_{10} \}$ and the A-Bot observing the same, plus the image $I$. Here, $c $ is the caption, and $q_{i},a_{i}$ is the $i^{th}$ dialog pair exchanged where $i={1\ldots 10}$. The action space for both bots consists of all possible output sequences of a specified maximum length (Q-Bot: 16, A-Bot: 9) under a fixed vocabulary (size 8645). Each action involves predicting words sequentially until a stop token is predicted, or the generated statement reaches the maximum length. Additionally, Q-Bot also produces a guess of the visual representation of the input image (VGG fc-7 embedding of size 4096). Both Q-Bot and A-Bot share the same objective and get the same reward to encourage cooperation. Information gain in each round of dialog is incentivized by setting the reward as the \textbf{change in distance} of the predicted image embedding to the ground-truth image representation. This means that a QA pair is of high quality only if it helps the Q-Bot make a better prediction of the image representation. Both policies are modeled by neural networks, as discussed in Section \ref{sec:arch}. 

A dialog round at time $t$ consists of the following steps: 1) the Q-Bot, conditioned on the state encoding, generates a question $q_{t}$, 2) A-Bot updates its state encoding with $q_{t}$ and then generates an answer $a_{t}$, 3) Both Q-Bot and A-Bot encode the completed exchange as a fact embedding, 4) Q-Bot updates its state encoding to incorporate this fact and finally 5) Q-Bot predicts the image representation for the unseen image conditioned on its updated state encoding.

Similar to Das et al. \cite{visdial}, we use the REINFORCE \cite{williams1992simple} algorithm that updates policy parameters in response to experienced rewards. The per-round rewards that are used to calculate the discounted returns follow:
\begin{equation}
r_{t}(s_{t}^{Q},(q_{t},a_{t},y_{t})) = l(y_{t-1},y^{gt}) - l(y_{t},y^{gt})
\label{eq:rewardfn}
\end{equation}
This reward is positive if the distance between image representation generated at time $t$ is closer to the ground truth than the representation at time $t-1$, hence incentivizing information gain at each round of dialog. The REINFORCE update rule ensures that a $(q_{t},a_{t})$ exchange that is informative has its probabilities pushed up. Do note that the image feature regression network $f$ is trained directly via supervised gradient updates on the L-2 loss.

\ak{However, as noted above, this RL optimization problem is ill-posed, since rewarding the agents for information exchange does not motivate them to stick to the rules and conventions of the English language. Thus, we follow an elaborate curriculum scheme described in \cite{visdial}. Specifically, for the first K rounds of dialog for each image, the agents are trained using supervision from the VisDial dataset. For the remaining 10-K rounds, however, they are trained via reinforcement learning. K starts at 9 and is linearly annealed to 0 over 10 epochs. Despite these modifications the bots are still observed to diverge from natural language and produce non-grammatical and incoherent dialog. Thus, we propose a multi bot architecture to help the agents interact in diverse and coherent, yet informative, dialog.}

\subsection{Multi-Agent Dialog Framework (MADF)}
\label{sec:madf}
\gs{In this section we describe our proposed Multi-Agent Dialog architecture in detail. We claim that if, instead of allowing a single pair of agents to interact, we were to make the agents more social, and make them \textit{interact and learn from multiple other agents}, they would be disincentivized to develop a private language, and would have to conform to the common language.} We call this Community Regularization.

\ak{In particular, we create either multiple Q-bots to interact with a single A-bot, or multiple A-bots to interact with a single Q-bot. All these agents are initialized with the learned parameters from the supervised pretraining as described in Section \ref{sec:supervised}. Then, for each batch of images from the VisDial dataset, we randomly choose a Q-bot to interact with the A-bot, or randomly choose an A-bot to interact with the Q-bot, as the case may be. The two chosen agents then have a complete dialog consisting of 10 question-answer pairs about each of those images, and update their respective weights based on the rewards received (as per Equation \ref{eq:rewardfn}) during the conversation, using the REINFORCE algorithm. This process is repeated for each batch of images, over the entire VisDial dataset. It is important to note that histories are \textit{not shared} across batches.} MADF can be understood in detail using the pseudocode in Algorithm \ref{alg:multibot}.

\begin{algorithm*}
\small
\caption{Multi-Agent Dialog Framework (MADF)}\label{alg:multibot}
\begin{algorithmic}[1]
\Procedure{MultiBotTrain}{}
\While{train\_iter < max\_train\_iter}\Comment{Main Training loop over batches}
    \State $Qbot \gets \textit{random\_select }(Q_1, Q_2, Q_3.... Q_q)$
    \State $Abot \gets \textit{random\_select }(A_1, A_2, A_3.... A_a)$\Comment{Either $q$ or $a$ is equal to 1}
    \State $history \gets (0,0,...0)$\Comment{History initialized with zeros}
    \State $fact \gets (0,0,...0)$\Comment{Fact encoding initialized with zeros}
    \State $\Delta image\_pred \gets 0$ \Comment{Tracks change in Image Embedding}
    \State $Qz_{1} \gets Ques\_enc(Qbot,fact,history,caption)$
    \For{t in 1:10}\Comment{Have 10 rounds of dialog}
        \State $ques_{t} \gets Ques\_gen(Qbot, Qz_{t})$
        \State $Az_{t} \gets Ans\_enc(Abot,fact,history,image,ques_{t},caption)$
        \State $ans_{t} \gets Ans\_gen(Abot, Az_{t})$
        \State $fact \gets [ques_{t}, ans_{t}]$ \Comment{Fact encoder stores the last dialog pair}
        \State $history \gets concat(history, fact)$ \Comment{History stores all previous dialog pairs}
        \State $Qz_{t} \gets Ques\_enc(Qbot,fact,history,caption)$
        \State $image\_pred \gets image\_regress(Qbot, fact, history, caption)$
        \State $ R_{t} \gets (target\_image - image\_pred)^2 - \Delta image\_pred$
        \State $\Delta image\_pred \gets (target\_image - image\_pred)^2$
    \EndFor
    \State $\Delta(w_{Qbot}) \gets \frac{1}{10}\sum_{t=1}^{10} \nabla_{\theta_{Qbot}} \left[G_t \log p(ques_{t},\theta_{Qbot}) - \Delta image\_pred \right]$ 
    \State $\Delta(w_{Abot}) \gets \frac{1}{10} \sum_{t=1}^{10} G_t \nabla_{\theta_{Abot}} \log p(ans_{t},\theta_{Abot}) $
    \State $w_{Qbot} \gets w_{Qbot} + \Delta(w_{Qbot})$ \Comment{REINFORCE and Image Loss update for Qbot}
    \State $w_{Abot} \gets w_{Abot} + \Delta(w_{Abot})$ \Comment{REINFORCE update for Abot}
    
\EndWhile\label{euclidendwhile}
\EndProcedure
\end{algorithmic}
\end{algorithm*}

\begin{table*}[t]
    \small
    \centering
    \caption{Comparison of answer retrieval metrics with previously published work}
    \begin{tabular}{|c|c|c|c|c|c|}
    \hline
    \textbf{Model} & \textbf{MRR} & \textbf{Mean Rank} & \textbf{R@10} \\
    \hline \hline
    Answer Prior \cite{visdial} & 0.3735 & 26.50 & 53.23 \\
    MN-QIH-G \cite{visdial} & 0.5259 & 17.06 & 68.88 \\
    HCIAE-G-DIS \cite{vdialog2} & 0.547 & 14.23 & 71.55\\
    Frozen-Q-Multi \cite{vdialog} & 0.437 & 21.13 & 60.48\\
    CoAtt-GAN \cite{vdialog3} & 0.5578 & 14.4 & 71.74\\
    \hline
    SL(Ours) & \textbf{0.610} & \textbf{5.323}  &\textbf{ 72.68}\\
    RL - 1Q,1A(Ours) & 0.459 & 7.097  & 72.34\\
    RL - 1Q,3A(Ours) & 0.601 & 5.495  & 72.48\\
    RL - 3Q,1A(Ours) & 0.590 & 5.56  & 72.61\\
    \hline
    \end{tabular}
    \label{tab:metrics}
\end{table*}

\section{Experiments and Results}

\subsection{Dataset description}
We use the VisDial 0.9 dataset for our task introduced by Das et al. \cite{visdial}. The data is collected using Amazon Mechanical Turk by pairing 2 annotators and asking them to chat about the image as a multi round VQA setup. One of the annotators acts as the questioner and has access to only the caption of the image and has to ask questions from the other annotator who acts as the `answerer' and must answer the questions based on the visual information from the actual image. This dialog repeats for 10 rounds at the end of which the questioner has to guess what the image was.
We perform our experiments on VisDial v0.9 (the latest available release) containing 83k dialogs on COCO-train and 40k on COCO-val images, for a total of 1.2M dialog question-answer pairs. We split the 83k into 82k for train, 1k for validation, and use the 40k as test, in a manner consistent with \cite{visdial}. The caption is considered to be the first round in the dialog history.

\begin{figure}
    \centering
    \includegraphics[width=0.4\textwidth]{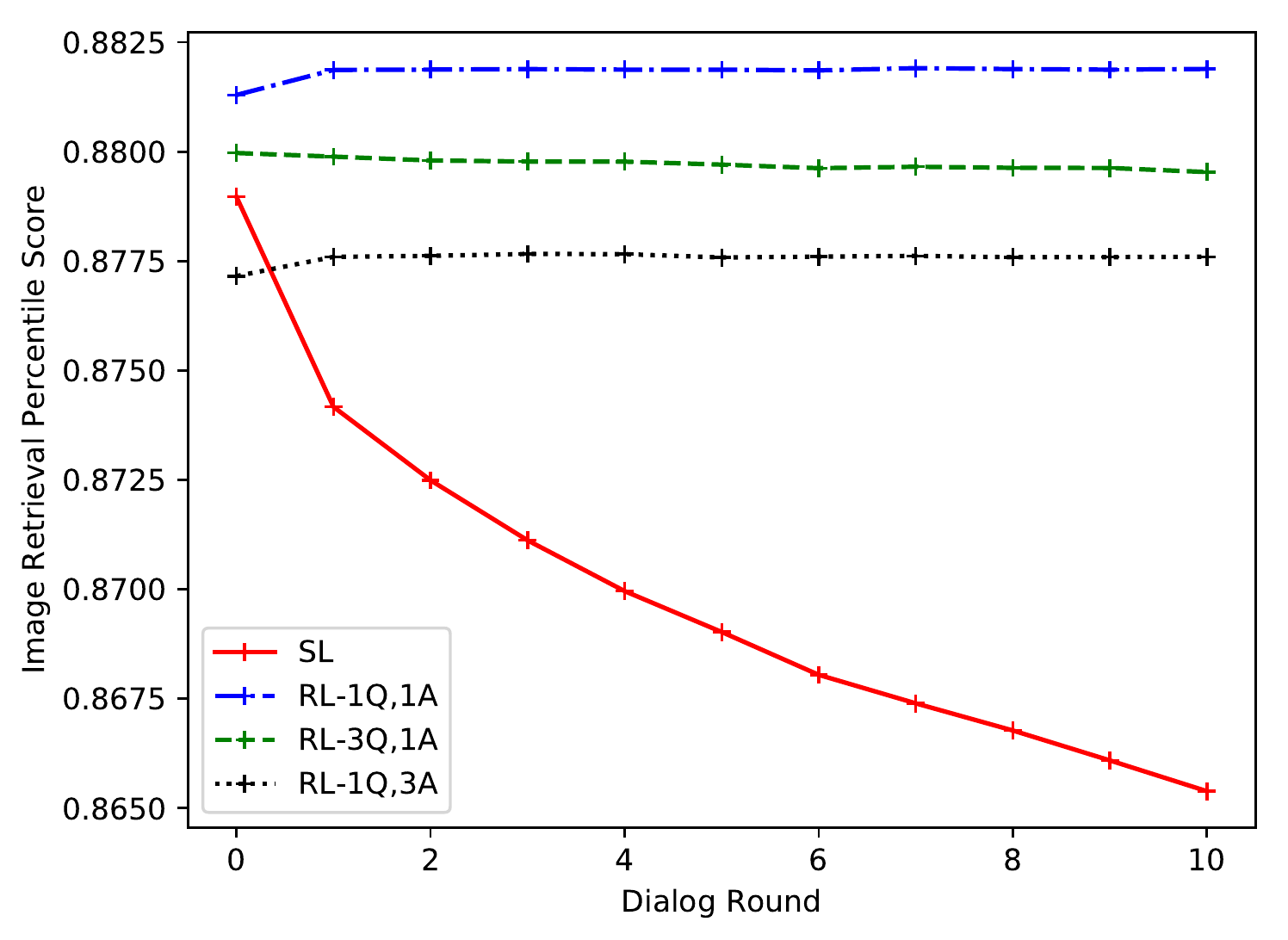}
    \caption{Comparison of Task Performance: Image Retrieval Percentile scores. This refers to the percentile scores of the ground truth image compared to the entire test set of 40k images, as ranked by distance from the Q-Bot's estimate of the image. The X-axis denotes the dialog round number (from 1 to 10), while the Y-axis denotes the image retrieval percentile score.}
    \label{fig:percentile}
\end{figure}

\begin{table*}
\centering
\caption{Human Evaluation Results - Mean Rank (Lower is better)}
\label{table:humaneval}
\begin{tabular}{|c|c|c|c|c|c|c|}
    \hline
    \textbf{} & \textbf{Metric} & \textbf{N} & \textbf{Supervised} & \textbf{RL 1Q,1A} & \textbf{RL 1Q,3A} & \textbf{RL 3Q,1A} \\
    \hline \hline
    1 & Question Relevance & 49 & \textbf{1.97} & 3.57 & 2.20 & 2.24 \\
    2 & Question Grammar & 49 & 2.16 & 3.67 & 2.24 & \textbf{1.91} \\
    3 & Overall Dialog Coherence: Q& 49 & 2.08 & 3.73 & 2.34 & \textbf{1.83} \\
    \hline
    4 & Answer Relevance & 53 & 2.09 & 3.77 & 2.28 & \textbf{1.84} \\
    5 & Answer Grammar & 53 & 2.20 & 3.75 & 2.05 & \textbf{1.98} \\
    6 & Overall Dialog Coherence: A & 53 & 2.09 & 3.64 & 2.35 & \textbf{1.90} \\
    
    \hline
\end{tabular}
\end{table*}

\subsection{Evaluation Metrics}
We evaluate the performance of our model's individual responses by using 4 metrics, proposed by \cite{vdialog}: \textbf{1) Mean Reciprocal Rank (MRR)}, \textbf{2) Mean Rank}, \textbf{3) Recall@10} and \textbf{4) Image Retrieval Percentile}. Mean Rank and MRR compute the average rank (and its reciprocal, respectively) assigned to the \akr{ground truth} answer, over a set of 100 candidate answers for each question (also averaged over all the 10 rounds).  Recall@10 computes the percentage of answers with rank less (better) than 10. Image Retrieval percentile is a measure of how close the image prediction generated by the Q-bot is to the ground truth. All the images in the test set are ranked according to their distance from the predicted image embedding, and the rank of the ground truth embedding is used to calculate the image retrieval percentile. \akr{All results for RL-1Q,1A, RL-1Q,3A and RL-3Q,1A are reported after 15 epochs of supervised learning and 10 epochs of curriculum learning as described in Section \ref{sec:training}. Consequently, the training time of all 3 systems are equal.}

\ak{Table \ref{tab:metrics} compares the Mean Rank, MRR, and Recall@10 of our agent architecture and dialog framework (below the horizontal line) with previously proposed architectures (above the line). SL refers to the agents after only the isolated, supervised training of Section \ref{sec:supervised}. RL-1Q,1A refers to a single, idiosyncratic pair of agents trained via reinforcement as in Section \ref{sec:rl}. RL-1Q,3A and RL-3Q,1A refer to social agents trained via our Multi-Agent Dialog framework in Section \ref{sec:madf}, with 1Q,3A referring to 1 Q-Bot and 3 A-Bots, and 3Q,1A referring to 3 Q-Bots and 1 A-Bot.

It can be seen that our agent architectures clearly outperform all previously published results using generative architectures in MRR, Mean Rank and R@10. This indicates that our approach produces consistently good answers (as measured by MRR, Mean Rank and  R@10). But it is important to note that the point here is not to demonstrate the superiority of our architecture compared to other architectures. The point here is instead to show that the MADF framework is able to recover the language quality of the supervised agent. In fact, community regularization (in the form of the proposed MADF setup) can be integrated with any of the visual dialog algorithms in Table \ref{tab:metrics}. 
Notice that SL has the best scores, which drops drastically in RL-1Q,1A. But the agents trained by MADF \akr{recover the scores obtained by SL. This shows that the agents trained by MADF are able to recover the language quality of SL agents without sacrificing much on the task performance (image retrieval percentile).} Fig. \ref{fig:percentile} shows the change in image retrieval percentile scores over dialog rounds. The percentile score decreases monotonically for SL, but is stable for all versions using RL.} \akr{The decrease in image retrieval score over dialog rounds for SL is because the test set questions and answers are not perfectly in-distribution (compared to the training set), and the SL system can't adapt to these samples as well as the systems trained with RL. As the dialog rounds increase, the out-of-distribution nature of dialog exchange increases, hence leading to a decrease in SL scores. Interestingly, despite having strictly more information in later rounds, the scores of RL agents do not increase - which we think is because of the nature of recurrent networks to forget. 

The results in Fig. \ref{fig:percentile} and Table \ref{tab:metrics} show that the MADF setup allows the agents to achieve consistent task performance without sacrificing on language quality. We further support this claim in the next section where we show that human evaluators rank the language quality of MADF agents to be much better than the agents trained via reinforcement without community regularization.}

\subsection{Human Evaluation}
\label{sec:humaneval}
There are no quantitative metrics to comprehensively evaluate dialog quality, hence we do a human evaluation of the generated dialog. There are 6 metrics we evaluate on: 1) Q-Bot Relevance, 2) Q-Bot Grammar, 3)A-Bot Relevance, 4) A-Bot Grammar, 5) Q-Bot Overall Dialog Coherence and 6) A-Bot Overall Dialog Coherence. We evaluate 4 Visual Dialog systems, trained via: 1) \textbf{Supervised Learning (SL)}, 2) \textbf{Reinforce for 1 Q-Bot, 1 A-Bot (RL-1Q,1A)}, 3) \textbf{Reinforce for 1 Q-Bot, 3 A-Bots (RL-1Q,3A)} and 4) \textbf{Reinforce for 3 Q-Bots, 1 A-Bot (RL-3Q,1A)}. We asked a total of 61 people to evaluate the 10 QA-pairs generated by each system for a total of 102 randomly chosen images, requiring them to give an ordinal ranking (from 1 to 4) for each metric. All the evaluators were provided with the caption from the dataset. 
Evaluators taking the perspective of the A-Bot were provided with the image and asked to evaluate answer relevance and grammar, while those taking the perspective of the Q-Bot evaluated question relevance and grammar. Both groups rated dialogs for overall coherence.
Table \ref{table:humaneval} contains the average ranks obtained on each metric (lower is better). 

\ak{The results convincingly validate our hypothesis that having multiple A-Bots/Q-Bots improves the language quality as compared with single Q-Bot and A-Bot. 
Kruskal-Wallis tests found strong differences among rankings (p< .0001) across all measures. Pairwise comparisons using the Mann-Whitney U test found a consistent pattern in which RL 1Q,1A performed substantially worse than other methods across all measures: for \textbf{Q-relevance}: SL: U=348, p<.0001; RL-1Q3A: U=2235, p< .0001; RL-3Q1A U=2209, p< .0001, \textbf{Q-grammar}: SL: U=319, p< .0001; RL-1Q3A U=2280, p < .0001; RL-3Q1A U=2221, p < .0001; \textbf{A-relevance}: SL U=256, p < .0001; RL-1Q3A U=2741, p < .0001; RL-3Q1A U-2909, p < .0001; \textbf{A-grammar}: SL U=305, p < .0001; RL-1Q3A U=2857, p < .0001; RL-3Q1A U=2673, p < .0001; \textbf{Overall (both groups)}: SL U=1206, p < .0001; RL-1Q3A U= 9458, p < .0001; RL-3Q1A U=10052, P < .0001.  Slight differences favoring RL 3Q,1A over RL 1Q,3A were found for A-relevance U=1889, p < .02 and overall coherence U=6543, p < .006 but otherwise SL, RL-1Q,3A, and RL-3Q,1A showed equivalent performance indicating that community regularization can effectively eliminate any losses to human intelligibility introduced through reinforcement learning.
These results further support the claims made in the previous section that the MADF setup allows the agents to show consistent task performance (image retrieval percentile) while maintaining the language quality of the supervised agents. 

We show a randomly chosen example from the set shown to the human evaluators in Fig. \ref{fig:humaneval}. The trends observed in the scores given by human evaluators are also clearly visible in this example. MADF agents are able to model the human responses much better than RL 1Q,1A and are about as well as (if not better) than SL trained agents. It can also be seen that although the RL-1Q,1A system has greater diversity in its responses, the quality of those responses is greatly degraded, with the A-Bot's answers especially being both non-grammatical and irrelevant.}

\section{Discussion and Conclusion}
\ak{In this paper we propose a novel community regularization technique, the Multi-Agent Dialog Framework (MADF), to improve the dialog quality of artificial agents. We show that training 2 agents with supervised learning does not ensure good task performance (measured by the image retrieval percentile scores) at test time, and it only deteriorates as the agents exchange more information about the image. We hypothesize that this is because the agents were trained in isolation and never allowed to interact during supervised learning, which leads to failure during testing when they encounter out of distribution samples (generated by the other agent, instead of ground truth) for the first time. We show how allowing a single pair of agents to interact and learn from each other via reinforcement learning dramatically improves their percentile scores, which additionally does not deteriorate over multiple rounds of dialog, since the agents have interacted with one another and been exposed to the other's generated questions or answers. However, in an attempt to improve task performance, the agents end up developing their own private language which does not adhere to the rules and conventions of human languages. As a result, the dialog system loses interpretability and sociability. To alleviate this issue, we propose a multi-agent dialog framework to provide regularization. In this framework, a single A-Bot is allowed to interact with multiple Q-Bots and vice versa. Through a human evaluation study, we show that this leads to significant improvements in dialog quality measured by relevance, grammar and overall coherence, \akr{without compromising the task performance}.}

\section{Future Work}
There are several possible extensions to this work. We plan to explore several other multi bot architectural settings and perform a more thorough human evaluation for qualitative analysis of our dialog. \ak{We also plan on incorporating MADF into other architectures and models proposed by more recent work and test how well MADF generalizes to other models.  }
Another avenue for future exploration is to use a richer image feature embedding to regress on. Currently, we use a regression network to compute the estimated image embedding which represents the Q-Bot's understanding of the image. However, a GAN which uses this embedding as a latent code to generate an image is an interesting possibility.

\section*{Acknowledgments} 
\ak{This research was sponsored in part by AFOSR Grant FA9550-15-1-0442. We would like to thank William Guss, Abhishek Das, Satwik Kottur and Fei Fang for their insightful and fruitful discussions and feedback. }

\bibliography{refs}
\bibliographystyle{aaai}

\end{document}